%% file: main.tex
\pdfoutput=1

\documentclass{article}
\usepackage{spconf,amsmath,graphicx}

\usepackage{multirow}
\usepackage{amssymb}
\usepackage{eucal}
\usepackage{booktabs}
\usepackage{tabularx} 
\usepackage{cite}
\usepackage{graphicx}
\usepackage{hyperref}

\usepackage{subcaption} 

\usepackage{color}
\usepackage{comment}
\usepackage{ifthen}

\usepackage{microtype}

\newcommand{\boldhdr}[1]{\noindent \textbf{#1.}}
\newcommand{\showcomments}{yes}
\newcommand\m[1]{
    \ifthenelse{\equal{\showcomments}{no}}{{\color{magenta} [Ming: #1]}}{\ignorespaces}
}
\newcommand{\kaiqi}[1]{\textcolor{black}{#1}}
\newcommand{\yt}[1]{\textcolor{black}{#1}}

\title{A Contrastive Knowledge Transfer Framework for Model Compression and Transfer Learning}
\name{Kaiqi Zhao, Yitao Chen, Ming Zhao\thanks{This work is partly supported by National Science Foundation awards CNS-1955593 and OAC-2126291. \kaiqi{Our code is at: \url{https://github.com/kaiqi123/CKTF.git}}.}}
\address{Arizona State University}

\begin{document}
%
\maketitle
%
\input{abstract}

\input{introduction}

\input{methodology}
\input{evaluation_setup}

\input{evaluation_model_compression}

\input{evaluation_transfer_learning}
\input{ablation}
\input{conclusion}

\vfill\pagebreak
\bibliographystyle{IEEEbib}
\bibliography{reference.bib}


\end{document}

%% file: abstract.tex
\begin{abstract}


Knowledge Transfer (KT) achieves competitive performance and is widely used \yt{for image classification tasks in model compression and transfer learning.}
Existing KT works transfer the information from a large model (``teacher'') to train a small model (``student'') by minimizing the difference of their conditionally independent output distributions.
However, these works overlook the high-dimension \textit{structural} knowledge from the intermediate representations of the teacher, which leads to limited effectiveness, and they are motivated by various heuristic intuitions, which makes it difficult to generalize.
This paper proposes a novel Contrastive Knowledge Transfer Framework (CKTF), which enables the transfer of sufficient structural knowledge from the teacher to the student by optimizing multiple \textit{contrastive objectives} across the intermediate representations between them.
Also, CKTF provides a \kaiqi{generalized agreement} to existing KT techniques and increases their performance significantly by deriving them as specific cases of CKTF.
The extensive evaluation shows that CKTF consistently outperforms the existing KT works by 0.04\% to 11.59\% in model compression and by 0.4\% to 4.75\% in transfer learning on various models and datasets.

\end{abstract}


%% file: introduction.tex
\vspace{-5pt}
\section{Introduction}\label{sec:introduction}
\vspace{-3pt}

Knowledge Transfer (KT) is an important and widely used technique for model compression and cross-domain transfer learning.
Deep neural networks (DNNs) are difficult to deploy on resource-constrained devices such as the Internet of Things (IoT) and smart devices~\cite{li2019edge}. KT can address this challenge by using the original model as the teacher to train a much smaller one as the student for deployment on edge devices.
Also, DNNs are difficult to train when there is insufficient labeled data.
KT can address this data deficiency by transferring knowledge from a teacher model in the source domain trained with abundant labeled data to the student model in the target domain where labels are unavailable.

Various KT techniques~\cite{hinton2015distilling,romero2014fitnets,zagoruyko2016paying,tung2019similarity,peng2019correlation,ahn2019variational,park2019relational,passalis2018probabilistic,heo2019knowledge,kim2018paraphrasing,yim2017gift,huang2017like} have been investigated for different image classification models.
Hinton et al. first introduced transferring soft logits (softmax outputs)~\cite{hinton2015distilling}, termed Knowledge Distillation (KD), by minimizing the KL divergence between the teacher's and student's soft logits and the \yt{cross-entropy} loss with the data labels. 
Later, other works~\cite{romero2014fitnets,zagoruyko2016paying,tung2019similarity,peng2019correlation,ahn2019variational,park2019relational,passalis2018probabilistic,heo2019knowledge,kim2018paraphrasing,yim2017gift,huang2017like} proposed to transfer various forms of intermediate representations, such as FSP matrix~\cite{yim2017gift} and attention~\cite{zagoruyko2016paying}.
%
However, these works assume that the output dimensions of intermediate layers are independent, and they let the student replicate the teacher's behavior by minimizing the difference between their probabilistic outputs. 
We argue that the intermediate representations are interdependent, and this minimization fails to capture the important structural knowledge of the teacher's convolution layers.
Also, the various KT works are motivated by different intuitions and lack a commonly agreed theory, which makes it challenging to generalize. 
Moreover, none of the existing KT works consistently outperform the conventional KD~\cite{hinton2015distilling}.


A recent work, CRD~\cite{tian2019contrastive} formulated KT as optimizing contrastive objectives, usually used for representation learning~\cite{van2018representation, gutmann2010noise, chen2020simple, tian2020makes}.
Their objective is to maximize a lower bound to the mutual information of the outputs of the penultimate layer (before soft logits) between the teacher and student~\cite{tian2019contrastive}.
However, the low dimensionality of the penultimate layer outputs restricts the amount of transferred information. 
Particularly in cross-domain transfer learning, the penultimate layer outputs of the teacher and student are irrelevant due to the extraneous data from different domains.
Moreover, the effectiveness of the contrastive objective on intermediate representations, which are high-dimension and crucial for guiding gradient updates, is currently unexplored.
%


To address the aforementioned limitations and improve the performance of KT for model compression and transfer learning, we propose a novel Contrastive Knowledge Transfer Framework (CKTF) to enable the transfer of sufficient structural knowledge from the teacher to the student by optimizing multiple \textit{contrastive objectives} across the intermediate representations between them.
CKTF defines \textit{positive representation pairs} as the outputs of the teacher's and student's intermediate modules from the same input sample and \textit{negative representation pairs} as from their modules' outputs given two different data samples, respectively. 
By optimizing the contrastive objectives constructed across all the modules, CKTF pushes each positive representation pair closer while pushing each negative representation pair farther apart, thereby achieving effective knowledge transfer.
Moreover, CKTF can incorporate and improve all the existing KT methods by adding their loss terms to the proposed contrastive loss during optimization. 


In this paper, we focus on applying CKTF for image classification models. Compared to the existing KT works, CKTF has several advantages:
%
first, compared to the output-level-only communication in the previous contrastive KT approach (CRD), CKTF allows the student to learn the intermediate layer state and to capture correlations of high-order output dependencies, leading to faster and better transfer; 
second, unlike the existing KT works which often perform worse than the conventional KD, CKTF consistently outperforms the conventional KD in all cases;
%
finally, CKTF provides a \kaiqi{generalized agreement} to existing KT methods and can incorporate existing works to significantly enhance their performance.



Our comprehensive evaluation shows that CKTF outperforms the existing KT works (KD, CRD, and 12 other solutions) significantly. 
For model compression using CIFAR-100 and Tiny-ImageNet, CKTF yields an accuracy improvement of 0.04\% to 11.59\% than the existing KT methods, and 0.95\% to 4.41\% compared to training the student directly using all the data. 
For transfer learning from Tiny-ImageNet to STL-10, CKTF converges faster than all the baselines and outperforms their accuracy by 0.4\% to 4.75\%.



%% file: methodology.tex
\vspace{-5pt}
\section{Methodology}\label{sec:methodology}
\vspace{-3pt}

\begin{figure}[t]
	\centering
    \includegraphics[width=\linewidth]{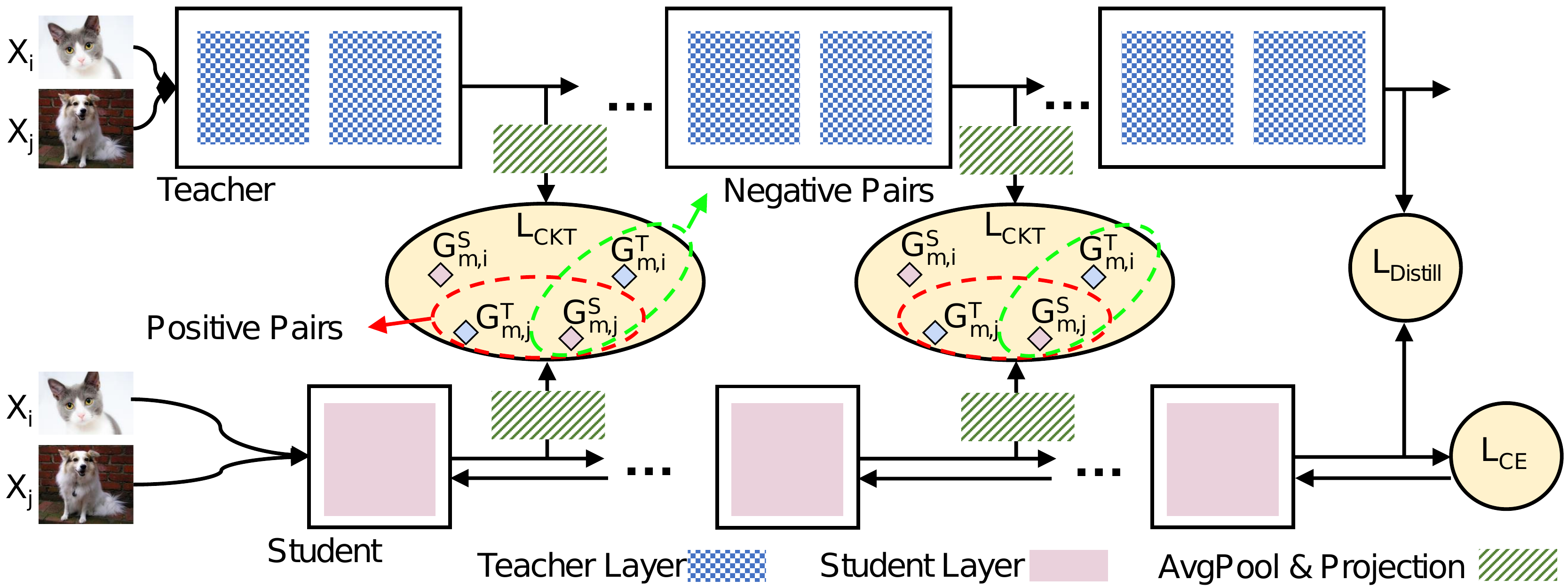}
    \vspace{-18pt}
	\caption{Workflow of CKTF.}
	\label{fig:workflow}
	\vspace{-15pt}
\end{figure}


\begin{table*}[t]
\scriptsize
\centering
\caption{\footnotesize{Top-1 test accuracy (\%) on CIFAR-100 and Tiny-ImageNet. Red/black arrows denote the increase/decrease compared to conventional KD.}}
\vspace{-10pt}
\label{tab:compression_overall_results}
\setlength{\tabcolsep}{4.7pt}
\begin{tabular*}{\textwidth}{l|ccccccc|cccc}
\toprule
DataSet           & \multicolumn{7}{c|}{CIFAR-100}                                                                                                                                                                                            & \multicolumn{4}{c}{Tiny-ImageNet}                                                                                         \\ \hline
Model             &                               &                               &                              &                               &                              &                              &                              &                              &                              &                              &                              \\
\hspace{2mm}Teacher           & WRN-40-2                      & WRN-40-2                      & ResNet-56                    & ResNet-110                    & ResNet-110                   & ResNet-32*4                  & VGG-13                       & VGG-19                       & VGG-16                       & ResNet-34                    & ResNet-50                    \\
\hspace{2mm}Student           & WRN-16-2                      & WRN-40-1                      & ResNet-20                    & ResNet-20                     & ResNet-32                    & ResNet-8*4                   & VGG-8                        & VGG-8                        & VGG-11                       & ResNet-10                    & ResNet-10                    \\
\hspace{2mm}Compression Ratio & 3.21                          & 3.96                          & 3.10                          & 6.24                          & 3.67                         & 6.03                         & 2.39                         & 5.01                         & 1.59                         & 4.28                         & 4.78                         \\ \hline
Baselines         &                               &                               &                              &                               &                              &                              &                              &                              &                              &                              &                              \\
\hspace{2mm}Teacher           & 75.61                         & 75.61                         & 72.34                        & 74.31                         & 74.31                        & 79.42                        & 74.64                        & 61.62                        & 61.35                        & 65.38                        & 65.34                        \\
\hspace{2mm}Student (w/o KT)    & 73.26                         & 73.54                         & 69.06                        & 69.06                         & 71.14                        & 72.5                         & 70.36                        & 54.61                        & 58.60                         & 58.01                        & 58.01                        \\ \hline
Method            &                               &                               &                              &                               &                              &                              &                              &                              &                              &                              &                              \\
\hspace{2mm}KD~\cite{hinton2015distilling}                & 74.92                         & 73.54                         & 70.66                        & 70.67                         & 73.08                        & 73.33                        & 72.98                        & 55.55                        & 62.51                        & 58.92                        & 58.63                        \\
\hspace{2mm}FitNet~\cite{romero2014fitnets}            & 73.58 ($\downarrow$)          & 72.24 ($\downarrow$)          & 69.21 ($\downarrow$)         & 68.99 ($\downarrow$)          & 71.06 ($\downarrow$)         & 73.50  ($\color{red} \uparrow$)           & 71.02 ($\downarrow$)         & 55.24 ($\downarrow$)         & 59.08 ($\downarrow$)         & 58.22 ($\downarrow$)         & 57.76 ($\downarrow$)         \\
\hspace{2mm}AT~\cite{zagoruyko2016paying}                & 74.08 ($\downarrow$)          & 72.77 ($\downarrow$)          & 70.55 ($\downarrow$)         & 70.22 ($\downarrow$)          & 72.31 ($\downarrow$)         & 73.44  ($\color{red} \uparrow$)          & 71.43 ($\downarrow$)         & 53.55 ($\downarrow$)         & 61.40 ($\downarrow$)          & 59.16  ($\color{red} \uparrow$)          & 58.92  ($\color{red} \uparrow$)          \\
\hspace{2mm}SP~\cite{tung2019similarity}                & 73.83 ($\downarrow$)          & 72.43 ($\downarrow$)          & 69.67 ($\downarrow$)         & 70.04 ($\downarrow$)          & 72.69 ($\downarrow$)         & 72.94 ($\downarrow$)         & 72.68 ($\downarrow$)         & 55.09 ($\downarrow$)         & 61.61 ($\downarrow$)         & 55.91 ($\downarrow$)         & 57.17 ($\downarrow$)         \\
\hspace{2mm}CC~\cite{peng2019correlation}                & 73.56 ($\downarrow$)          & 72.21 ($\downarrow$)          & 69.63 ($\downarrow$)         & 69.48 ($\downarrow$)          & 71.48 ($\downarrow$)         & 72.97 ($\downarrow$)         & 70.71 ($\downarrow$)         & 54.87 ($\downarrow$)         & 58.34 ($\downarrow$)         & 57.18 ($\downarrow$)         & 57.36 ($\downarrow$)         \\
\hspace{2mm}VID~\cite{ahn2019variational}               & 74.11 ($\downarrow$)          & 73.3 ($\downarrow$)           & 70.38 ($\downarrow$)         & 70.16 ($\downarrow$)          & 72.61 ($\downarrow$)         & 73.09 ($\downarrow$)         & 71.23 ($\downarrow$)         & 54.94 ($\downarrow$)         & 60.07 ($\downarrow$)         & 58.53 ($\downarrow$)         & 57.65 ($\downarrow$)         \\
\hspace{2mm}RKD~\cite{park2019relational}               & 73.35 ($\downarrow$)          & 72.22 ($\downarrow$)          & 69.61 ($\downarrow$)         & 69.25 ($\downarrow$)          & 71.82 ($\downarrow$)         & 71.90 ($\downarrow$)          & 71.48 ($\downarrow$)         & 54.13 ($\downarrow$)         & 59.96 ($\downarrow$)         & 57.35 ($\downarrow$)         & 57.05 ($\downarrow$)         \\
\hspace{2mm}PKT~\cite{passalis2018probabilistic}               & 74.54 ($\downarrow$)          & 73.45 ($\downarrow$)          & 70.34 ($\downarrow$)         & 70.25 ($\downarrow$)          & 72.61 ($\downarrow$)         & 73.64  ($\color{red} \uparrow$)          & 72.88 ($\downarrow$)         & 55.35 ($\downarrow$)         & 60.46 ($\downarrow$)         & 58.41 ($\downarrow$)         & 58.66  ($\color{red} \uparrow$)          \\
\hspace{2mm}AB~\cite{heo2019knowledge}                & 72.50 ($\downarrow$)           & 72.38 ($\downarrow$)          & 69.47 ($\downarrow$)         & 69.53 ($\downarrow$)          & 70.98 ($\downarrow$)         & 73.17 ($\downarrow$)         & 70.94 ($\downarrow$)         & 50.31 ($\downarrow$)         & 55.65 ($\downarrow$)         & 57.22 ($\downarrow$)         & 58.05 ($\downarrow$)         \\
\hspace{2mm}FT~\cite{kim2018paraphrasing}                & 73.25 ($\downarrow$)          & 71.59 ($\downarrow$)          & 69.84 ($\downarrow$)         & 70.22 ($\downarrow$)          & 72.37 ($\downarrow$)         & 72.86 ($\downarrow$)         & 70.58 ($\downarrow$)         & 53.65 ($\downarrow$)         & 58.84 ($\downarrow$)         & 56.22 ($\downarrow$)         & 56.48 ($\downarrow$)         \\
\hspace{2mm}FSP~\cite{yim2017gift}               & 72.91 ($\downarrow$)          & N/A                           & 69.95 ($\downarrow$)         & 70.11 ($\downarrow$)          & 71.89 ($\downarrow$)         & 72.62 ($\downarrow$)         & 70.23 ($\downarrow$)         & N/A                          & N/A                          & N/A                          & N/A                          \\
\hspace{2mm}NST~\cite{huang2017like}               & 73.68 ($\downarrow$) & 72.24 ($\downarrow$) & 69.60 ($\downarrow$) & 69.53 ($\downarrow$) & 71.96 ($\downarrow$)         & 73.30 ($\downarrow$)          & 71.53 ($\downarrow$)         & 51.08 ($\downarrow$)         & 58.47 ($\downarrow$)         & 59.23  ($\color{red} \uparrow$)          &  47.83 ($\downarrow$)                           \\ \hline
\hspace{2mm}CRD~\cite{tian2019contrastive}               & 75.48  ($\color{red} \uparrow$)           & 74.14  ($\color{red} \uparrow$)           & 71.16  ($\color{red} \uparrow$)          & 71.46  ($\color{red} \uparrow$)           & 73.48  ($\color{red} \uparrow$)          & 75.51  ($\color{red} \uparrow$)          & 73.94  ($\color{red} \uparrow$)          & 56.99  ($\color{red} \uparrow$)          & 62.04 ($\downarrow$)         & 60.02  ($\color{red} \uparrow$)          & 59.31  ($\color{red} \uparrow$)          \\
\hspace{2mm}CKTF              & \textbf{75.85  ($\color{red} \uparrow$)}  & \textbf{74.49  ($\color{red} \uparrow$)}  & \textbf{71.20  ($\color{red} \uparrow$)}  & \textbf{71.80  ($\color{red} \uparrow$)}   & \textbf{73.84  ($\color{red} \uparrow$)} & \textbf{75.74  ($\color{red} \uparrow$)} & \textbf{74.31  ($\color{red} \uparrow$)} & \textbf{57.57  ($\color{red} \uparrow$)} & \textbf{63.01  ($\color{red} \uparrow$)} & \textbf{60.39  ($\color{red} \uparrow$)} & \textbf{59.42  ($\color{red} \uparrow$)} \\ \hline
\hspace{2mm}CRD+KD~\cite{tian2019contrastive}            & 75.64  ($\color{red} \uparrow$)           & 74.38  ($\color{red} \uparrow$)           & 71.63  ($\color{red} \uparrow$)          & 71.56  ($\color{red} \uparrow$)           & 73.75  ($\color{red} \uparrow$)          & 75.46  ($\color{red} \uparrow$)          & 74.29  ($\color{red} \uparrow$)          & 58.09  ($\color{red} \uparrow$)          & 63.66  ($\color{red} \uparrow$)          & 61.99  ($\color{red} \uparrow$)          & 61.26  ($\color{red} \uparrow$)          \\
\hspace{2mm}CKTF+KD            & \textbf{75.89  ($\color{red} \uparrow$)}  & \textbf{74.94  ($\color{red} \uparrow$)}  & \textbf{71.86  ($\color{red} \uparrow$)} & \textbf{71.66  ($\color{red} \uparrow$)}  & \textbf{74.07  ($\color{red} \uparrow$)} & \textbf{75.97  ($\color{red} \uparrow$)} & \textbf{74.55  ($\color{red} \uparrow$)} & \textbf{58.76  ($\color{red} \uparrow$)} & \textbf{63.97  ($\color{red} \uparrow$)} & \textbf{62.31  ($\color{red} \uparrow$)} & \textbf{61.51  ($\color{red} \uparrow$)} \\
\bottomrule
\end{tabular*}
\end{table*}

\begin{table*}[t]
\vspace{-5pt}
\scriptsize
\centering
\caption{\footnotesize{Top-1 test accuracy (\%) of KT methods incorporated into CKTF. Numbers inside the parentheses denote the improvement over the original method.}}
\vspace{-10pt}
\label{tab:compression_ours_on_others}
\begin{tabular*}{\textwidth}{l|cccccccccc}
\toprule
                                                                                                    & CKTF+FitNet                                                     & CKTF+AT                                                         & CKTF+SP                                                         & CKTF+CC                                                         & CKTF+VID                                                        & CKTF+RKD                                                        & CKTF+PKT                                                        & CKTF+AB                                                         & CKTF+FT                                                         & CKTF+NST                                                        \\ \hline
\begin{tabular}[c]{@{}l@{}}T: ResNet-32$\times$4\\ S: ResNet-32$\times$4\\ (CIFAR-100)\end{tabular} & \begin{tabular}[c]{@{}l@{}}73.18\\ (1.68 $\color{red} \uparrow$)\end{tabular} & \begin{tabular}[c]{@{}l@{}}74.92\\ (1.48 $\color{red} \uparrow$)\end{tabular} & \begin{tabular}[c]{@{}l@{}}75.30\\ (2.36 $\color{red} \uparrow$)\end{tabular} & \begin{tabular}[c]{@{}l@{}}75.86\\ (2.89 $\color{red} \uparrow$)\end{tabular} & \begin{tabular}[c]{@{}l@{}}75.43\\ (2.34 $\color{red} \uparrow$)\end{tabular} & \begin{tabular}[c]{@{}l@{}}74.92\\ (3.02 $\color{red} \uparrow$)\end{tabular} & \begin{tabular}[c]{@{}l@{}}75.82\\ (2.18 $\color{red} \uparrow$)\end{tabular} & \begin{tabular}[c]{@{}l@{}}75.38\\ (2.21 $\color{red} \uparrow$)\end{tabular} & \begin{tabular}[c]{@{}l@{}}75.39\\ (2.53 $\color{red} \uparrow$)\end{tabular} & \begin{tabular}[c]{@{}l@{}}75.08\\ (1.78 $\color{red} \uparrow$)\end{tabular} \\ \hline
\begin{tabular}[c]{@{}l@{}}T: VGG-19\\ S: VGG-8\\ (Tiny-ImageNet)\end{tabular}                      & \begin{tabular}[c]{@{}l@{}}56.19\\ (0.95 $\color{red} \uparrow$)\end{tabular} & \begin{tabular}[c]{@{}l@{}}55.33\\ (1.78 $\color{red} \uparrow$)\end{tabular} & \begin{tabular}[c]{@{}l@{}}56.22\\ (1.13 $\color{red} \uparrow$)\end{tabular} & \begin{tabular}[c]{@{}l@{}}55.99\\ (1.12 $\color{red} \uparrow$)\end{tabular} & \begin{tabular}[c]{@{}l@{}}56.34\\ (1.4 $\color{red} \uparrow$)\end{tabular}  & \begin{tabular}[c]{@{}l@{}}55.96\\ (1.83 $\color{red} \uparrow$)\end{tabular} & \begin{tabular}[c]{@{}l@{}}56.82\\ (1.47 $\color{red} \uparrow$)\end{tabular} &                       \begin{tabular}[c]{@{}l@{}}52.63\\(2.32 $\color{red} \uparrow$)\end{tabular} & \begin{tabular}[c]{@{}l@{}}56.39\\(2.74 $\color{red} \uparrow$)\end{tabular} & \begin{tabular}[c]{@{}l@{}}51.97\\ (0.89$\color{red} \uparrow$)\end{tabular} \\ 
\bottomrule
\end{tabular*}
\end{table*}

\subsection{Framework Overview}
\vspace{-3pt}

Let $X = \{x_i\}_{i=1}^B$ and $Y = \{y_i\}_{i=1}^B$ denote a set of inputs with a batch size of $B$ and its ground truth labels $Y$, respectively.
We define a module as a group of convolution layers. 
The output representations of the modules from the teacher and student can be described as $\{T_m\}_{m=1}^M$ and $\{S_m\}_{m=1}^M$, respectively, where $M$ denotes the number of modules.
%
Similarly, let $T_h$ and $S_h$ denote the output vectors of the penultimate layer from the teacher and student, respectively.

Figure~\ref{fig:workflow} illustrates the workflow of the proposed Contrastive Knowledge Transfer Framework (CKTF). 
The optimization objective in CKTF consists of three components: 
1) the cross entropy loss with the ground truth labels;
2) the proposed contrastive loss to transfer knowledge from the intermediate representations $\{T_m\}_{m=1}^M$ and the penultimate layer $T_h$ of the teacher to $\{S_m\}_{m=1}^M$ and $S_h$ of the student, respectively;
and 3) the distillation loss from other KT methods.
The loss function of CKTF is as follows:
\begin{equation}\label{equ:overall_loss}
    \small
    \begin{split}
     	L = \gamma L_{CE} (Y, S_h) +& L_{CKT} (\{T_m\}_{m=1}^M, \{S_m\}_{m=1}^M, T_h, S_h) \\
                                   +& \theta L_{Distill} (T_h, S_h)
    \end{split}
\end{equation}
where $\gamma$ equals either 1 or 0 depending on the availability of labels, and $\theta$ is a hyper-parameter that controls the weight of the loss term.

The first loss term in Eq.~\ref{equ:overall_loss} enforces the supervised learning from labels, which is typically implemented as a \yt{cross-entropy} loss for classification tasks: $L_{CE} (Y, S_h) = \sum_{i=1}^{c} [Y_{i}log(S_{h,i}) + (1 - Y_i)log(1 - S_{h,i})]$, where $c$ denotes the number of classes of the dataset. 
%
The second loss term is the proposed contrastive loss that transfers \kaiqi{high-dimension structural knowledge} from both the intermediate presentations and the penultimate layer via contrastive learning. 
\kaiqi{It works because, as opposed to just transferring information about conditionally independent output class probabilities, the multiple contrastive objectives constructed in $L_{CKT}$ better transfer all the information in the teacher's representational space (see Section~\ref{sec:contrastive_kt} for details).}
The third loss term is used to incorporate existing KT methods into CKTF.
For example, for the conventional KD~\cite{hinton2015distilling}, it is defined as the KL-divergence-based loss that minimizes the difference between the teacher's and student's soft logits: $L_{Distill} (T_h, S_h) = KL (softmax(T_h/\rho) || softmax(S_h/\rho))$, where $\rho$ is the temperature. 
In this way, CKTF can help improve the performance of existing KT methods (see Section~\ref{sec:eva_compression} for evaluation results).

Note that, in transfer learning, $\gamma$ (Eq.~\ref{equ:overall_loss}) is set to zero since supervision from labels is not available.





\vspace{-5pt}
\subsection{Contrastive Knowledge Transfer}\label{sec:contrastive_kt}
\vspace{-3pt}



CKTF constructs the contrastive loss across intermediate representations from multiple modules of the teacher and student. 
Directly using intermediate representations $\{T_m\}_{m=1}^M$ and $\{S_m\}_{m=1}^M$ to perform contrastive learning is infeasible, since
1) the dimension between $T_m$ and $S_m$ might be different, 
and 2) the huge feature dimension of $T_m$ and $S_m$ may cause memory issues or significantly increase the training time. 
In detail, the dimension of $S_m$ is calculated as $|S_m| = B \times o_m^s \times (k_m^s)^2$, where $B$, $o_m^s$ and $k_m^s$ denote the batch size, output dimension, and kernel size of the module $m$ of the student. 
For example, for ResNet-50 on Tiny-ImageNet with a batch size of 32, the feature dimension of one intermediate module can be: $32 \times 1024 \times 16^2 \approx 8.39$ millions, and its teacher may also have a similar level of the feature dimension.

To solve the above problem, CKTF first applies an average pooling over $T_m \in R^{B \times o_m^t\times k_m^t\times k_m^t}$ and $S_m \in R^{B \times o_m^s \times k_m^s \times k_m^s}$, respectively, and it produces the output $\bar{T}_m \in R^{B\times o_m^t \times 1 \times 1}$ and $\bar{S}_m \in R^{B\times o_m^s \times 1 \times 1}$, respectively. 
Then it uses a reshaping function $h(\cdot)$ that changes the 4-D $\bar{T}_m$ and $\bar{S}_m$ to a 2-D space, yielding $H^T_m \in R^{B*o^t_m}$ and $H^S_m \in R^{B*o^s_m}$, respectively:
\begin{equation}
    \vspace{-3pt}
    \begin{split}
        &\bar{S}_m = AvgPool(S_m), \bar{T}_m = AvgPool(T_m) \\
        &H_m^S = h(\bar{S}_m), H_m^T = h(\bar{T}_m)
    \end{split}
    \vspace{-3pt}
\end{equation}

Next, a projection network $g(\cdot)$ takes the presentations $\{H_m^T\}_{m=1}^M$ and $\{H_m^S\}_{m=1}^M$ as the input, and for the module $m$, it produces: $G^T_m=g(H_m^T) \in R^{B \times d}$ and $G^S_m=g(H_m^T) \in R^{B \times d}$, respectively, where $d$ denotes the output dimension of the projection network.
%
$g(\cdot)$ used in CKTF is a single linear layer of size $d=128$ followed by the $\ell_2$ normalization.
Note that $g(\cdot)$ is discarded after training, so we do not change the model architecture.
We will show that the linear projection is better than Multi-Layer Perceptron (MLP) projection used in representation learning~\cite{van2018representation, gutmann2010noise, chen2020simple, tian2020makes} and discuss the effect of $d$ in Section~\ref{sec:ablation}.

%

CKTF constructs the contrastive loss using $\{G^T_m\}_{m=1}^M$ and $\{G^S_m\}_{m=1}^M$.
Given a batch of random samples $X = \{x_i\}_{i=1}^B$, we define \textit{positive representation pairs} as $(G_{m,i}^S, G_{m,i}^T)$, which are the outputs of the student's and teacher's module $m$ from the same input sample $x_i$, and \textit{negative representation pairs} as $(G_{m,i}^S, G_{m,j}^T)$ from their modules' outputs given two different data samples $x_i$ and $x_j$, respectively.

%

CKTF aims to push closer each positive pair $G_{m,i}^S$ and $G_{m,i}^T$ for every input $x_i$, while pushing $G_{m,i}^S$ apart from $\{G_{m,j}^T\}_{j=1, j \neq i}^N$.
$N$ is the number of negative representation pairs.
CKTF defines the contrastive loss based on the intermediate representations as follows:
\begin{equation}
\vspace{-1pt}
    \small
	L_{MCKT} (G_{m}^S, G_{m}^T) = - E \left[ log\frac{f(G_{m,i}^S, G_{m,i}^T)}{\sum_{j=1}^{N}f(G_{m,i}^S, G_{m,j}^T)} \right] 
 \vspace{-0.5pt}
\end{equation}
where the function $f(\cdot)$ is similar with that used in \cite{van2018representation, gutmann2010noise, chen2020simple, tian2020makes}, specifically, $f(G_{m,i}^S, G_{m,i}^T) = \frac{ exp(G_{m,i}^S G_{m,i}^T /\tau) }{ exp(G_{m,i}^S G_{m,i}^T /\tau) + N/N_d }$.
$N_d$ is the number of training samples of the dataset, and $\tau$ is a temperature that controls the concentration level. 
The previous works~\cite{van2018representation, gutmann2010noise, chen2020simple, tian2020makes} use the function for different domains or objectives, such as self-supervised representation learning~\cite{chen2020simple} and density estimation~\cite{gutmann2010noise}, whereas we are the first to construct multiple contrastive objectives on the intermediate representations of image classification models for knowledge transfer. 
The minimization of the contrastive loss $L_{MCKT}$ is maximizing the lower bound of the mutual information~\cite{van2018representation, gutmann2010noise, chen2020simple, tian2020makes} between $\{G_{m}^T\}_{m=1}^M$ and $\{G_{m}^S\}_{m=1}^M$.

Similar to $L_{MCKT}$, CKTF constructs the contrastive objective on the outputs of the penultimate layer as:
\begin{equation}
\vspace{-1pt}
    \small
    L_{PCKT} (S_h, T_h) = - E \left[ log\frac{f(S_{h,i}, T_{h,i})}{\sum_{j=1}^{N}f(S_{h,i}, T_{h,j})} \right]
\vspace{-0.5pt}
\end{equation}
Finally, the proposed contrastive loss (the second loss term in Eq.~\ref{equ:overall_loss}) is defined as the weighted sum of $L_{MCKT}$ and $L_{PCKT}$: 
\begin{equation}
    \small
	L_{CKT} = \alpha_1 \sum_{m=1}^{M} L_{MCKT} (G_{m}^S, G_{m}^T) + \alpha_2 L_{PCKT}(T_h, S_h).
\end{equation}

%% file: evaluation_setup.tex
\section{Evaluation}\label{sec:evaluation}
\vspace{-3pt}

\begin{figure}[t]
  \centering
  \begin{subfigure}{.305\linewidth}
    \centering
    \includegraphics[width=2.8cm]{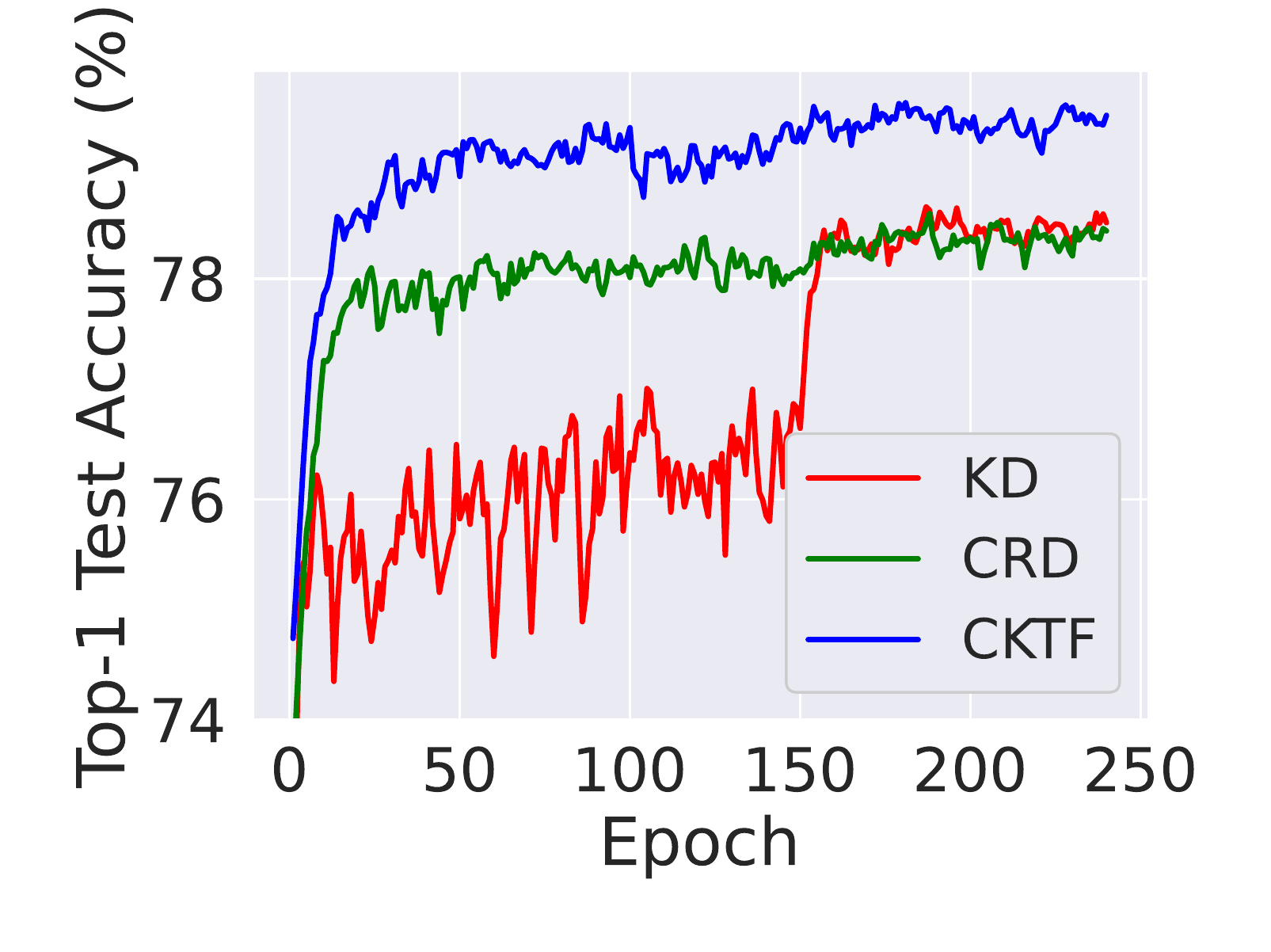}
    \vspace{-20pt}
    \caption{\scriptsize{T:VGG-19/S:VGG-19}}
  \end{subfigure}
  \begin{subfigure}{.305\linewidth}
    \centering
    \includegraphics[width=2.8cm]{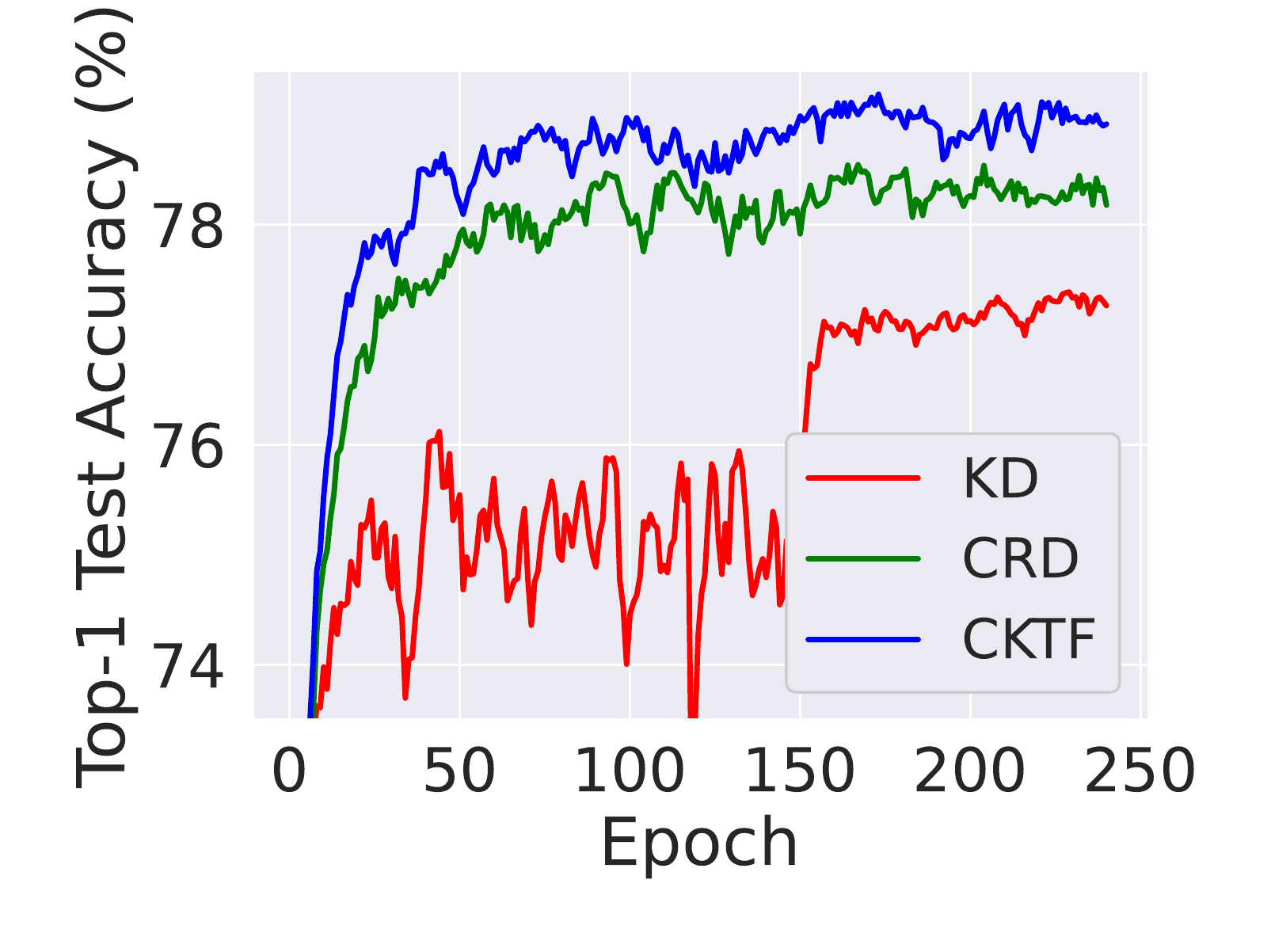}
    \vspace{-20pt}
    \caption{\scriptsize{T:VGG-19/S:VGG-8}}
  \end{subfigure}
  \begin{subfigure}{.35\linewidth}
    \centering
    \includegraphics[width=2.8cm]{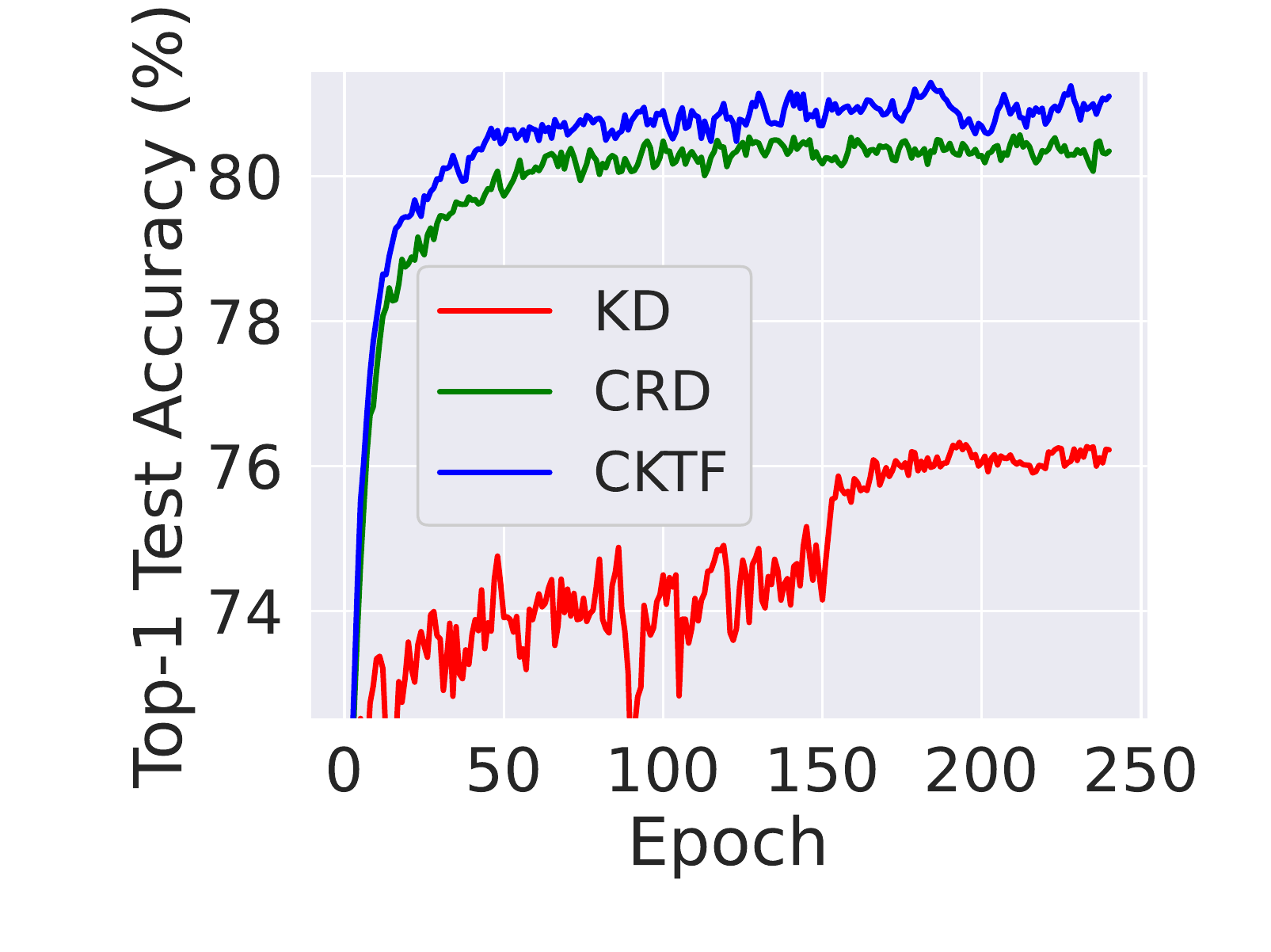}
    \vspace{-8pt}
    \caption{\scriptsize{T:ResNet-18/S:ResNet-18}}
  \end{subfigure}
  \vspace{-5pt}
  \caption{\footnotesize{Top-1 test accuracy of KD, CRD, and the proposed CKTF on STL-10 when transferring knowledge from Tiny-ImageNet.}}
   \label{fig:transfer_tinyimagenet_stl10}
  \vspace{-10pt}
\end{figure}

\begin{figure}[t]
  \centering
  \begin{subfigure}{.45\linewidth}
    \centering
    \includegraphics[width=4.0cm]{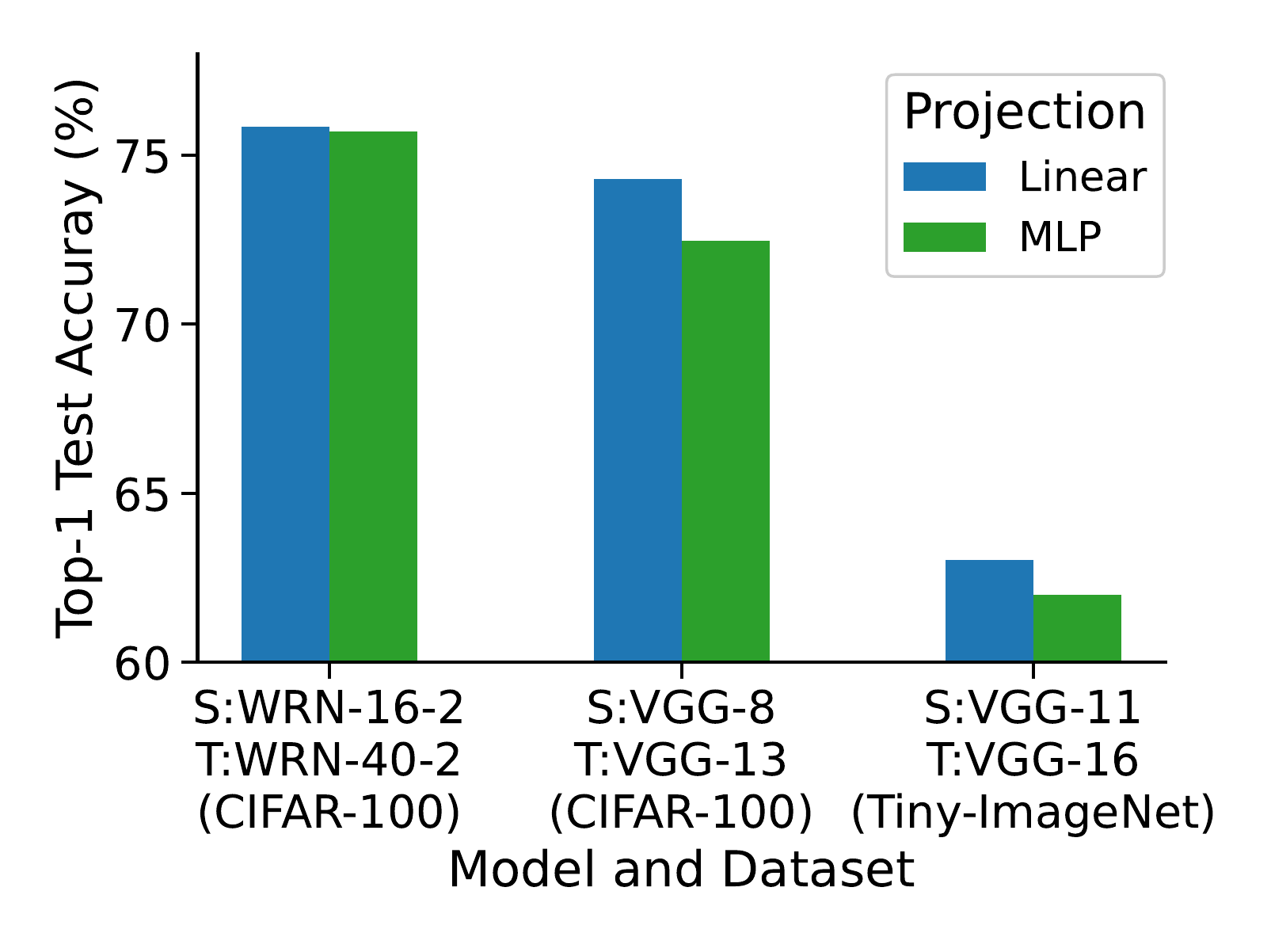}
    \vspace{-20pt}
    \caption{\footnotesize{Projection}}
    \label{fig:ablation_head}
  \end{subfigure}
  \begin{subfigure}{.45\linewidth}
    \centering
    \includegraphics[width=4.0cm]{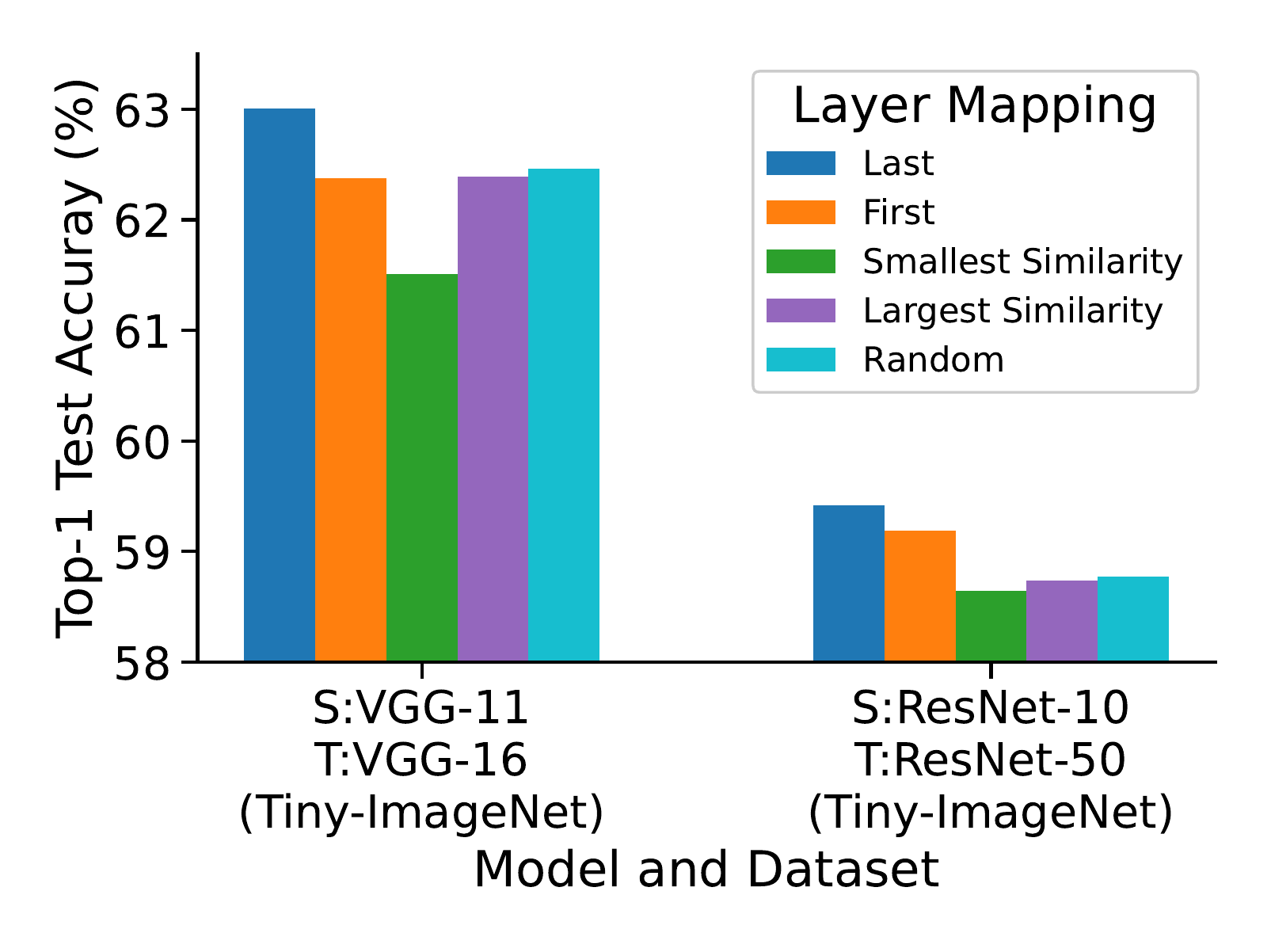}
    \vspace{-20pt}
    \caption{\footnotesize{Layer Mapping}}
    \label{fig:ablation_layer_selection}
  \end{subfigure}
  \vspace{-8pt}
  \caption{\footnotesize{The effect of the (a) projection type and (b) layer mapping.}}
  \vspace{-5pt}
\end{figure}

\smallskip
\boldhdr{Models and Datasets}
We conducted an extensive evaluation of Wide ResNet Networks (WRN), ResNet, and VGG models on 
1) CIFAR-100~\cite{krizhevsky2009learning} that consists of 50K 32$\times$32 RGB images with 100 classes, 
2) Tiny-ImageNet~\cite{le2015tiny} containing 100K 64$\times$64 images with 200 classes, 
and 3) STL-10~\cite{coates2011analysis} that contains 5K 10-category labeled training images, 8K test images, and 100K unlabeled images.

\smallskip
\boldhdr{Implementation Details} 
We implemented CKTF on PyTorch version 1.9.0 and conducted experiments on 4 \kaiqi{Nvidia RTX 2080 Ti GPUs}.
The learning rate is initialized to 5e-2 and decays with a rate of 0.1 at epochs 150, 180, and 210.
The total training epochs is 240.
Weight decay is set to 5e-4. 
Nesterov SGD optimizer is used with a momentum of 0.9. 
$N$ and $\tau$ are set to 16384 and 0.1, respectively.
When CKTF is evaluated alone, $\theta$ is 0, and $\alpha 1$ and $\alpha 2$ are set to 0.8 and 0.2, respectively.
When evaluating the related KT methods incorporated into CKTF, \kaiqi{$\theta$ is set to 1, which means the third loss term in Eq.~\ref{equ:overall_loss} is the same as the loss used in their papers}, and $\alpha_1$ and $\alpha_2$ follow the above settings (0.8 and 0.2).

%% file: evaluation_model_compression.tex
\vspace{-5pt}
\subsection{Model Compression}\label{sec:eva_compression}
\vspace{-3pt}

%
We compare CKTF for model compression tasks with three baselines: 
1) a large model that is uncompressed and directly trained (teacher), 
2) a small model directly trained without KT (student w/o KT),
and 3) the same small model trained with various KT methods.
For the implementation, we use the public CRD code-base~\cite{tian2019contrastive} to conduct a fair comparison.

Table~\ref{tab:compression_overall_results} presents the Top-1 test accuracy of various teacher and student combinations on CIFAR-100 and Tiny-ImageNet. 
CFKT significantly outperforms all the existing KT methods in all cases. Specifically, CFKT outperforms:
1) the conventional KD~\cite{hinton2015distilling} by 0.5\% to 2.41\% (none of the existing methods consistently outperforms KD),
2) the other KT methods by 0.04\% to 11.59\%,
and 3) the related contrastive learning method CRD~\cite{tian2019contrastive} (the second best in the results) by 0.04\% to 0.97\%.

Compared to the student trained directly on the data without KT, CFKT is 0.95\% to 4.41\% better.
\kaiqi{We also observe that, compared to the student w/o KT, CFKT performs better on Tiny-ImageNet (better than the student w/o KT by 4.41\% to 1.41\%) than on CIFAR-100 (better than the student w/o KT by 3.95\% to 0.95\%). 
This could be because Tiny-ImageNet is more complicated than CIFAR-100 (with more classes and data), resulting in more complicated intermediate representation, whereas CFKT is good at capturing this complicated high-dimension structural knowledge.}
Further, CFKT enables the small student to achieve comparable performance to the large teacher with only 16\% of its original size. 
This confirms that CFKT is beneficial to on-device image classification applications that require small, high-performance models. 


\smallskip
\boldhdr{Results on Incorporating KT Methods}
We measure the performance of existing KT methods incorporated into CFKT (following Eq.~\ref{equ:overall_loss}), using ResNet-32×4/ResNet-8×4 and VGG-19/VGG-8 as the teacher/student on CIFAR-100 and Tiny-ImageNet.
As shown in Table~\ref{tab:compression_ours_on_others}, the Top-1 test accuracy of the existing KT works is significantly improved by 0.89\% to 3.02\% when incorporated into CFKT.  
The results demonstrate that CFKT provides a \kaiqi{generalized agreement} behind knowledge transfer.
\kaiqi{Another observation is that, when incorporating existing KT methods into CFKT, the improvement on the methods that transfer from the last several layers is higher than the methods that transfer from intermediate representations. 
For example, PKT+CKTF and SP+CKTF achieve an improvement of 2.18\% and 2.36\%, compared to PKT and SP, respectively, whereas AT+CKTF and FitNet+CKTF achieve an improvement of 1.48\% and 1.68\%, compared to AT and FitNet, respectively. 
This is because methods that transfer from the last several layers lack the teacher's intermediate information, which can be compensated by CFKT after they are incorporated into CFKT. So the improvement is larger.
For the methods that transfer knowledge from intermediate representations, the transferred information is partial since they do not explicitly capture correlations or higher-order dependencies in representations.
The integration of CKTF though still provides additional intermediate information, the improvement to the final accuracy is smaller than that from the methods completely lacking intermediate information.
}


%% file: evaluation_transfer_learning.tex
\vspace{-5pt}
\subsection{Transfer Learning}
\vspace{-3pt}


%
We first train the teacher using the source domain data (Tiny-ImageNet) with ground truth labels. 
Then, we transfer knowledge from the teacher to the student using the unlabeled data from the target domain (STL-10) with KT methods. 
Finally, we \kaiqi{fine-tune} the student (only train its linear classifier) using the training set of STL-10 and evaluate its accuracy on the test set of SLT-10.
This is a common practice~\cite{alain2016understanding, zhang2017split, tian2019contrastive} for evaluating the quality of transfer learning.

We compare CKTF with KD~\cite{hinton2015distilling} and CRD~\cite{tian2019contrastive}.
The teacher and student can be either the same, e.g., VGG-19/VGG-19, or different, e.g., VGG-19/VGG-8.
Figure~\ref{fig:transfer_tinyimagenet_stl10} shows how the Top-1 test accuracy of the student evolves during fine-tuning on STL-10.
CKTF converges faster than all the baselines and outperforms them in final accuracy by 0.4\% to 4.75\%.
This result validates that CKTF is advantageous for cross-domain transfer learning, even without labeled data in the target domain.


%% file: ablation.tex
\vspace{-5pt}
\subsection{Ablation Study}\label{sec:ablation}
\vspace{-3pt}



\smallskip
\boldhdr{Effect of Projection}
Figure~\ref{fig:ablation_head} compares the Top-1 test accuracy of three student models trained with CKTF, using the linear vs. MLP projection network (discussed in Section~\ref{sec:contrastive_kt}), on CIFAR-10 and Tiny-ImageNet.
Linear projection outperforms MLP projection by 0.15\% to 1.85\%. 



\smallskip
\boldhdr{Effect of Output Dimension}
We analyze the impact of the output dimension $d$ of the linear projection network on four student models by varying the value of $d$ from 16 to 128. 
We observe that a larger output always leads to better performance, and $d=128$ is better than others by 0.03\% to 2.2\% on CIFAR-10 and Tiny-ImageNet.

\smallskip
\boldhdr{Effect of Layer Mapping}
Figure~\ref{fig:ablation_layer_selection} illustrates the effect of five strategies for mapping the teacher's and student's layers in each module, including mapping the student's last layer with the teacher's 1) first, 2) last, or 3) a randomly chosen convolution layer or mapping between the layer pair whose outputs have the 4) largest or 5) smallest cosine similarity.
Last-layer mapping outperforms others by 0.23\% to 1.5\% on CIFAR-10 and Tiny-ImageNet.

%% file: conclusion.tex
\section{Conclusions}\label{sec:conclusions}
\vspace{-3pt}


This paper proposes a novel Contrastive Knowledge Transfer Framework (CKTF) for model compression and transfer learning in image classification.
Different from previous KT works, 
CKTF enables the transfer of high-dimension structural knowledge between the teacher and student by optimizing multiple contrastive objectives across the intermediate representations. 
It also provides a \kaiqi{generalized agreement} to existing KT methods and increases their accuracy significantly by deriving them as specific cases of CKTF.
An extensive evaluation shows that CKTF consistently outperforms the existing KT works by 0.04\% to 11.59\% in model compression and by 0.4\% to 4.75\% in transfer learning.
\kaiqi{In the future, we will investigate the effectiveness of CKTF in ensemble knowledge transfer and large-scale language model compression.}


